\documentclass{article}

\PassOptionsToPackage{numbers}{natbib}

\usepackage[preprint]{neurips_2023}

\usepackage[utf8]{inputenc} 
\usepackage[T1]{fontenc}    
\usepackage{hyperref}       
\usepackage{url}            
\usepackage{booktabs}       
\usepackage{amsfonts}       
\usepackage{nicefrac}       
\usepackage{microtype}      
\usepackage{xcolor}         
\usepackage{graphicx}
\usepackage{multirow}
\usepackage{array}
\usepackage{tabularx}
\usepackage{algorithm}
\usepackage{algorithmicx}
\usepackage{algpseudocode}
\usepackage{amsmath}
\usepackage{listings}
\usepackage{amsthm}
\usepackage{setspace}

\newcolumntype{Y}{>{\centering\arraybackslash}X}

\newcolumntype{Y}{>{\centering\arraybackslash}X}

\title{SHARP: Sparsity and Hidden Activation RePlay for Neuro-Inspired Continual Learning}

\author{%
  Mustafa Burak Gurbuz\\
  School of Computer Science \\
  Georgia Institute of Technology, USA \\
  \texttt{mgurbuz6@gatech.edu} \\
  \And
  Jean Michael Moorman \\
  School of Computer Science \\
  Georgia Institute of Technology, USA \\
  \texttt{jmoorman9@gatech.edu} \\
  \And
  Constantine Dovrolis \\
  School of Computer Science \\
  Georgia Institute of Technology, USA \\
  The Cyprus Institute, Cyprus \\
  \texttt{constantine@gatech.edu}
}

\begin{document}

\maketitle

\begin{abstract}
Deep neural networks (DNNs) struggle to learn in dynamic environments since they rely on fixed datasets or stationary environments. Continual learning (CL) aims to address this limitation and enable DNNs to accumulate knowledge incrementally, similar to human learning. Inspired by how our brain consolidates memories, a powerful strategy in CL is replay, which involves training the DNN on a mixture of new and all seen classes. However, existing replay methods overlook two crucial aspects of biological replay: 1) the brain replays processed neural patterns instead of raw input, and 2) it prioritizes the replay of recently learned information rather than revisiting all past experiences. To address these differences, we propose SHARP, an efficient neuro-inspired CL method that leverages sparse dynamic connectivity and activation replay. Unlike other activation replay methods, which assume layers not subjected to replay have been pretrained and fixed, SHARP can continually update all layers. Also, SHARP is unique in that it only needs to replay few recently seen classes instead of all past classes. Our experiments on five datasets demonstrate that SHARP outperforms state-of-the-art replay methods in class incremental learning. Furthermore, we showcase SHARP's flexibility in a novel CL scenario where the boundaries between learning episodes are blurry. The SHARP code is available
at \url{https://github.com/BurakGurbuz97/SHARP-Continual-Learning}.
\end{abstract}

\section{Introduction}

Deep Neural Networks (DNNs) are usually trained on static data distributions. In this process, training batches are created by either drawing samples from a shuffled and fixed dataset or by gathering observations from an unchanging environment. While this approach has led to remarkable progress across a wide range of domains \cite{alphaGo, deepLearning, visualUnderstanting}, DNNs miss a crucial aspect of general intelligence: the ability to continuously learn over time in a dynamic environment. In such a Continual Learning (CL) scenario, DNNs overwrite previously acquired knowledge when learning from new data, resulting in a challenging phenomenon called Catastrophic Forgetting (CF) \cite{McCloskey}.

Unlike DNNs, animals show a remarkable ability to retain and integrate knowledge obtained in ever-changing environments \cite{embracing_change, Neuroscience-Inspired}. This inspires us to scrutinize some mechanisms in the brain to alleviate CF in DNNs. Although even simpler animals like \textit{Drosophila} \cite{drosophila} and \textit{Caenorhabditis elegans} \cite{c_elagans, c.elegans2} also exhibit some CL, we focus on the mammalian brain.

\textbf{Activation replay \hypertarget{insight:1}{(Insight-1)}: } Replay of neural patterns is a potent mechanism that the brain leverages to learn new information without forgetting existing knowledge \cite{gido2020, hayes2021replay}. Similarly, many CL approaches use replay by combining new examples with previous ones and training the network with this mixture \cite{parisi2018_survey, delange_survey}. However, this replay is usually done by replaying raw samples (e.g., pixel-level images), which is not biologically plausible. In contrast, replay in the hippocampus follows spatial trajectories rather than raw sensory inputs \cite{Karlsson2009}, and it is time-compressed compared to experience \cite{Dupret2010}, pointing to the brain replaying highly processed representations.

{\centering
    \emph{(1) Drawing inspiration from the brain, we replay hidden layer activations to reduce forgetting.}
\par}

\textbf{Replay consolidates recently learned information \hypertarget{insight:2}{(Insight-2)}: }Place cells are a classification of excitatory neurons found in the hippocampus that fire when an animal is at a specific location in an environment. Replay recruits place cells in the same sequence as during prior experience, allowing for experience-dependent memory consolidation. Research on memory replay in the brain has found correlations between the firing patterns of place cells during sleep and those observed while performing recent waking tasks \cite{recent_replay1, recent_replay2, hayes2021replay}. This replay of recent learning has been confirmed in follow-up studies and has also been observed in brain areas beyond the hippocampus \cite{Louie2001, Davidson2009, ji2007, Peyrache2009}.

Furthermore, the temporal dynamics of memory replay involve prioritizing the replay of recent information, which gradually diminishes as the replay of new information replaces it \cite{Carr2011, O'Neill2010, recent_replay_decay, Karlsson2009}. This is in contrast with current practice in CL. To our knowledge, all existing replay-based CL methods involve revisiting all past experiences uniformly while learning a new task \cite{delange_survey, parisi2018_survey, hayes2021replay}. For instance, in image classification, a typical replay strategy involves creating training batches that include examples not only from novel classes but also from all previously encountered classes.

{\centering
    \hypertarget{inspiration_2}{\emph{(2) Similar to replay in the brain, we replay a limited number of recently learned classes.}}
\par}

\textbf{Sparse and dynamic connectivity \hypertarget{insight:3}{(Insight-3)}: }DNNs typically have fully connected layers. Dense 
 connectivity within these layers is susceptible to forgetting since changes in a single connection impact the activations of all subsequent units. In contrast, the brain relies on sparse and dynamic connectivity (e.g., synaptic pruning and formation). For instance, anatomical studies reveal that cortical pyramidal neurons receive relatively few excitatory inputs from neighboring neurons \cite{Hunter2021TwoSA, sparse_conn_pyramidal1, sparse_conn_pyramidal2}. Furthermore, after synaptic pruning during early childhood, the connection density in the brain remains relatively constant in healthy adults \cite{synaptic_pruning_development}. In other words, the brain does not sacrifice sparsity to learn; instead, it employs synaptic  plasticity to form more effective neural pathways. 

{\centering
    \emph{(3) Inspired from the brain's sparse connectivity and synaptic  plasticity, we begin with a sparse DNN and maintain the network's density while rewiring connections throughout the CL process.}
\par}

\textbf{Synaptic stability \hypertarget{insight:4}{(Insight-4)}: }
Another mechanism the brain uses to preserve past knowledge is the stabilization of dendritic spines (single synaptic inputs). For example, experiments on mice show that while learning new skills, the volume of some neuronal dendritic spines increases. Following the learning window, these enlarged dendritic spines persist, despite the subsequent learning of other tasks \cite{yang2009}. Conversely, when these spines are experimentally removed, the mouse forgets the associated skill \cite{cichon2015, hayashi2015}. This implies that stable spines play a vital role in long-term information storage. 

{\centering
    \emph{(4) Like stable spines, we freeze input connections of specific hidden units to alleviate forgetting.}
\par}

In summary, our main contributions in this paper are as follows:
\begin{itemize}
    \item We propose a novel CL method that employs dynamic connectivity and replays activations of recently seen classes to achieve state-of-the-art performance in class incremental learning.
    \item Prior activation replay methods use pretrained/fixed feature extractors. Thanks to sparse dynamic connectivity, our method is the first instance of activation replay without pretraining.
    \item We demonstrate the flexibility of our method on a new type of CL scenario in which the boundaries between different learning episodes are blurred, and episodes may overlap.
\end{itemize}

\section{Related Work}
Replay serves as an effective strategy for addressing CF and is widely used in the CL literature \cite{delange_survey, parisi2018_survey, hayes2021replay}. Current replay approaches in the literature can be grouped into three categories.

\textbf{Raw Example Replay: }The most prevalent form of replay involves preserving and revisiting previous examples \cite{DER, EReplay, icarl, Caccia2022, fdr}. These methods employ a memory with a fixed budget and fill it by sampling from encountered examples. To approximate static data distributions, the training batches are formed by mixing  examples from the input stream with a chosen set of older examples from memory.

Although the replay of raw examples helps to counteract forgetting, it has several drawbacks.  Firstly, this type of replay is biologically implausible (see \hyperlink{insight:1}{Insight-1}). One may argue that biological plausibility is not a significant concern from a practical standpoint. However, raw replay also has practical limitations. For instance, it is not suitable for real-world applications where previous user data cannot be stored indefinitely. Additionally, raw replay is often criticized because storing a portion of all observed examples oversimplifies the problem. For example, a recent study \cite{GDumb} introduced the GDumb algorithm, which stores a subset of samples in memory as they arrive and, during testing, trains a model from scratch using only the samples in memory. Surprisingly, this straightforward baseline surpasses almost all raw replay methods. Hence, storing and retraining on raw images approximates static data rather than providing models with the ability to adapt to shifting data distributions.

\textbf{Generative Replay: }Some studies have suggested using a generative model to  produce samples that resemble previously encountered examples \cite{DGR, PseudoRecursal, Lifelong-generative-modeling}. These methods tackle concerns about user data rights and may also be backed by neuroscience, as the hippocampus plays a generative role in memory consolidation. Nevertheless, these methods bring a considerable computational overhead. Also, training the generative model can be challenging since it is also vulnerable to forgetting and has additional issues, such as the quality of generated samples degrading over time \cite{gido2020}.

\textbf{Activation Replay: }Few neuro-inspired works have suggested addressing the limitations of raw or generated example replay by replaying activations. This is in line with \hyperlink{insight:1}{Insight-1}.  For instance, Brain-inspired replay (BIR) \cite{gido2020} models the interplay between the neocortex and hippocampus, providing biologically plausible generative activations replay to mitigate forgetting. Similarly, REMIND \cite{remind} replays compressed activations and is inspired by hippocampal indexing theory \cite{hippo_indexing}.

However, these methods assume layers not receiving replay are pretrained and frozen. For instance, REMIND pretrains the first 15 convolutional layers of the Resnet-18 and initiates replay after the first 15 frozen layers. This makes activation replay practically the same as raw replay since the mapping between raw images and stored activations never changes. Moreover, these methods fail without pretraining and freezing since stored activations become outdated when earlier layers receive gradient updates. So, rather than tackling CF, they conceal it by freezing all layers that do not receive replay.

Lastly, existing replay methods revisit all past classes to create batches resembling a static dataset. Thus, replay in CL is insensitive to input novelty, which differs greatly from how the brain replays experiences since it prioritizes replaying novel experiences. Our approach addresses all mentioned limitations of replay methods. First, we replay activations without pretraining or fully freezing feature extraction layers. Second, we only replay few recently seen classes. Third, by frequent connection rewiring, we break the input-output paths that cause interference and promote the formation of more efficient paths that facilitate positive knowledge transfer across different learning experiences.

\textbf{Other Approaches in CL: }Replay is not the only strategy to mitigate forgetting. Regularization methods modulate weight updates to protect the important parameters preserving past knowledge \cite{EWC, SI, MAS, HAT, RWALK}. However, they perform poorly compared to even the most simple replay methods \cite{scenarios}. On the other hand, parameter isolation methods assign different parameters for every new task to reduce forgetting. This separation is often accomplished by growing new branches \cite{PNN, DEN, DeepAdaptation} or partitioning existing connections \cite{neural_pruning, adaptive_group_sparse, nispa, space_net}. These approaches are mainly designed for scenarios in which a task identifier is given during testing to exclude irrelevant network parts. Without task identifiers, they either fail or significantly underperform compared to replay methods \cite{scenarios}.

\textbf{SHARP and Parameter Isolation Methods: }Our method is designed for a more demanding scenario in which models make predictions without task information. However, it shares some similarities with parameter isolation approaches, particularly with NISPA \cite{nispa}. NISPA begins with a pruned network and rewires connections while maintaining the same network density. Additionally, it freezes certain connections. Our approach adopts a similar strategy to rewire units. Thus, both NISPA and our method share \hyperlink{insight:3}{Insight-3} and \hyperlink{insight:4}{Insight-4}. It is crucial to note, however, that NISPA requires task identifiers during evaluation to mask certain output units and fails in our scenario.

\begin{figure}
  \centering
  \includegraphics[width=0.80\textwidth]{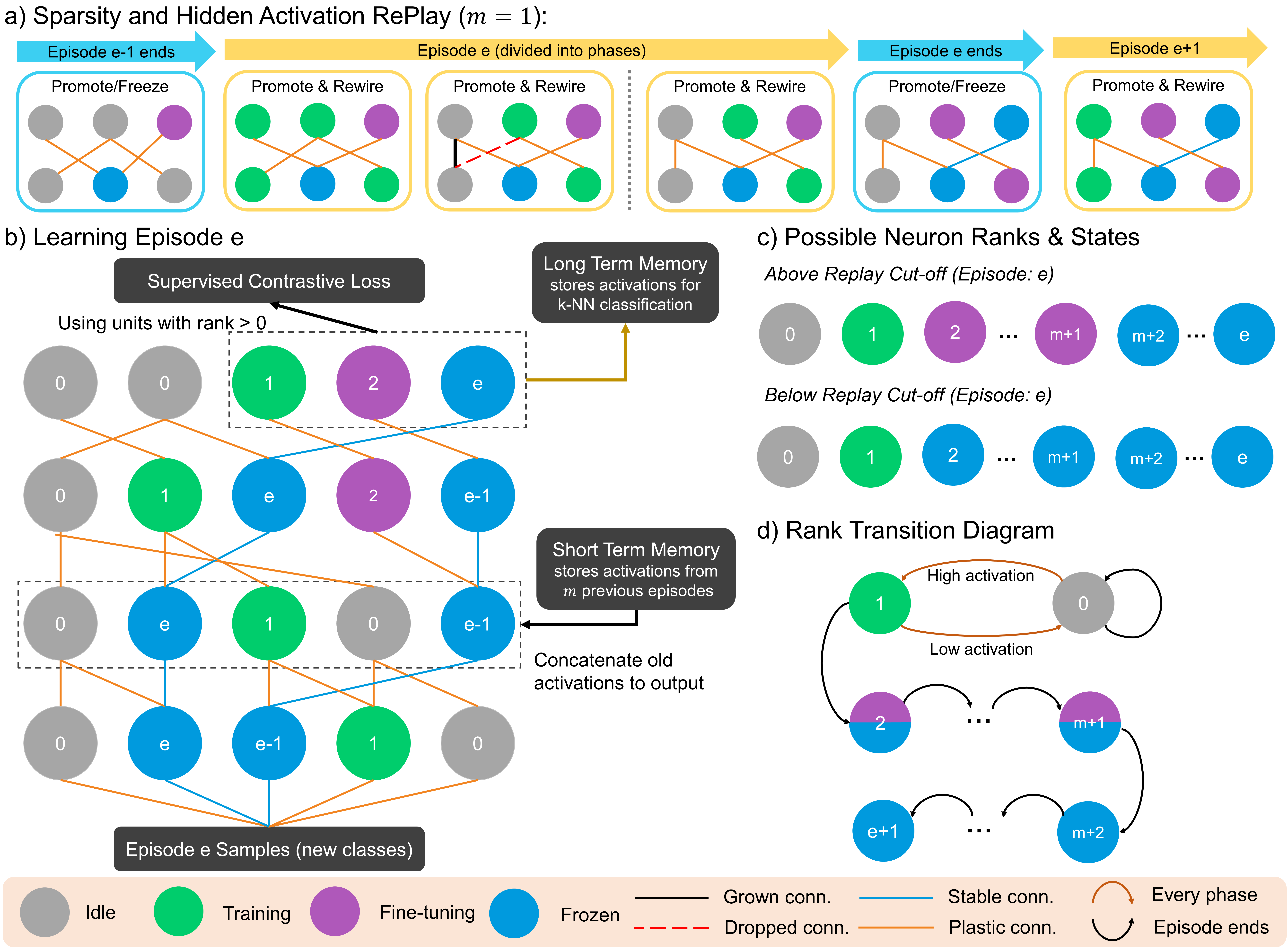}
  \caption{a) Shows how SHARP  unfolds over time. b) Depicts the network structure with a toy example. c) Displays possible unit ranks and states. d) Represents the diagram of rank transitions.}
  \label{fig:main_figure}
  \vskip -0.15in
\end{figure}

\section{SHARP: Sparsity and Hidden Activation RePlay}
Our CL approach SHARP, \textbf{S}parsity and \textbf{H}idden \textbf{A}ctivation \textbf{R}e\textbf{P}lay, sequentially processes learning episodes. We break down each episode into ``phases'' that last for a fixed number of epochs (see \textbf{Figure}~\ref{fig:main_figure}-a). Before each phase, we rewire units based on local activations. The rewiring aims to reduce negative interference among units while forming new paths. During a phase, SHARP employs Supervised Contrastive Loss \cite{SUPCON} and replays hidden activations stored in short-term memory (see \textbf{Figure}~\ref{fig:main_figure}-b). At the end of an episode, we stabilize few units and create long-term representations that are used for making predictions. See Supplementary Material for complete pseudocode.

\subsection{Problem Formulation and Notation}
We focus on class incremental learning (CIL) through a series of $E$ episodes where each episode $e$ has a training dataset $D_e$ \cite{scenarios}. Although the term ``task'' is commonly used in the CL literature to describe these episodes, we avoid using it due to its frequent use in different contexts. We partition our DNN with $L$ layers into two modules. The first module $G(\cdot)$ comprises the initial $K$ layers. It takes an input image $x_i$ and outputs activation  $h_i$. The second module, $F(\cdot)$ has remaining $L - K$ layers, and it maps $h_i$ to the final layer activation $z_i$. Furthermore, our DNN has a fixed connection density $d$. Here, $d$ is the ratio of connections remaining after pruning compared to before pruning. Pruning is done randomly at initialization and applied separately to each layer, ensuring every layer has the same density $d$. In convolutional layers, a ``unit'' is replaced by a 3-d convolution filter. Likewise, a ``connection'' is replaced by a 2-d  kernel.  All units are followed by ReLU.

SHARP utilizes a Short-Term Memory (STM) that temporarily holds $k$ activations from $m$ previous episodes. Unless specified otherwise, we set $m=1$, meaning STM stores $k$ activations solely from the most recent episode. The STM follows three simple rules: (1) It evenly distributes the capacity $k$ among stored classes. (2) It randomly selects samples from the input stream to be stored. (3) It employs linear quantization to represent a floating-point value using a single byte.

\subsection{Unit Ranks}
We assign each unit a rank from $0$ to $e$, where $e$ is the number of episodes seen. We denote units at layer $l$ with $U^l$ and use subscripts to denote units with certain ranks (e.g., $U^{l}_{r < 2}$). At the initialization, all units are rank-0. Before each phase, we promote some units from rank $0$ to $1$, as detailed below. Once an episode ends, the ranks of all units with a rank $>0$ are incremented by one (see \textbf{Figure}~\ref{fig:main_figure}-d).

\textbf{Rank transition before every phase: } We assign rank-1 to some units based on activations. This step resembles NISPA's ``candidate stable'' unit selection, but NISPA is not suitable for CIL \cite{nispa}. To do so, first,  we compute the activation at each layer on some episode examples (1024 for experiments):
\vskip -0.03in
\begin{equation}
    A_l =  \sum_{x \sim D_e} a_l(x) =  \sum_{x \sim D_e} \sum_{u_i^l \in U^{l}}   a_{u_i^l}(x)
\end{equation}

Here, $a_{u_i^l}(x)$ is the activation of unit $i$ at layer $l$ for sample $x$ and $a_l(x)$ is the cumulative activation of layer $l$ for sample $x$. Next, we select rank-1 units denoted by set $P^l_{1}$ as follows:

\begin{equation}
     \underset{P^l_{1} \subseteq  U^{l}_{r < 2}}{\min} |P^l_{1}| \; \; \text{subject to}\; \; \sum_{x \sim D_e}   \sum_{u_i^l \in P^l_{1}} a_{u_i^l}(x) \geq \tau  A_l  - \sum_{u_i^l \in U^{l}_{r \geq 2}} a_{u_i^l}(x)   
\end{equation}

All $U^{l}_{r < 2}$ units that are not selected are demoted to rank-0. Our aim is to find the smallest set of units to mark as rank-1 to capture at least a fraction $\tau$ (explained shortly) of the total activation in layer $l$, excluding the activations generated by units in $U^{l}_{r \geq 2}$. If units with ranks $\geq 2$ capture most of the activation, only few units need to be chosen. Thus, high-rank units contribute to satisfying the constraint without increasing the minimization objective. We solve this problem using a greedy algorithm. First, we sort units based on activations in descending order. Second, we select units with the highest activation until the target activation is achieved or exceeded. This greedy algorithm is optimal for this problem, as we show in the Supplementary Material. Note that input units always have the highest possible rank (i.e., $E$) by definition and do not participate in the selection.

\textbf{Calculating $\tau$: } As discussed in \cite{nispa}, in the initial phases of training, the activations may exhibit unpredictable fluctuations. Thus, we should start with a larger $\tau$. As training progresses, the unit activations become more stable. So, we can then tighten the selection criterion. This reasoning implies that we should gradually decrease $\tau$, beginning with large $\tau_{1}$ in phase-1 and reducing $\tau$ in increments that grow larger with each phase. We achieve this using a cosine annealing schedule: $\tau_p = \max(\tau_{min}, \frac{1}{2} \left ( 1 + \cos \left ( \frac{p+1}{k} \times \pi \right ) \right ))$. Here, $p$ is the phase index, and $k$ (set to 30) determines the shape of the function. Also, we do not allow $\tau_p$ to be smaller than $\tau_{min}$ which is a hyperparameter.

\subsection{Connection Rewiring: }
Our first rewiring rule is to drop the connections from all rank-0 units to rank-1 units after each $P^l_{1}$ selection. Since the ranks of $U^{l}_{r > 0}$ units are incremented by one after each episode, we ensure that there will be no connections from lower-rank to higher-rank units. We can rephrase this property as:

{\centering
    \emph{\textbf{Path Property: } There is no directed path in DNN from rank-$i$ unit to rank-$j$ unit if $i < j$.}
\par}

Note that each unit in a DNN represents a function that is characterized by its input connections and the input connections of all ancestor units. So, weight updates can alter the functionality of a unit in two ways, directly by changing the weights of incoming connections into the unit and second, by changing incoming connections of ancestor units. Combining this observation with  \emph{Path Property}:

{\centering
    \emph{\textbf{Observation: } Considering $i < j$, we can update any rank-$i$ unit without impacting the functionality of any rank-$j$ unit, as there are no paths connecting units with rank-$i$ to those with rank-$j$.}
\par}

After dropping some connections from layer $l$, we add the same amount of new connections to layer $l$, ensuring the density $d$ of each layer remains constant. We should not break \emph{Path Property} because the corresponding \emph{Observation} is crucial for avoiding forgetting. Thus, we have the following constraint:

{\centering
    \emph{\textbf{Growing Constraint: } We cannot grow a connection from rank-$i$ to rank-$j$ unit if $i < j$. }
\par}

So, we can connect rank-$i$ units to rank-$j$ units if $i \geq j$. We have observed that putting all the connection quota into increasing incoming connections to rank-0 units yields better performance. A potential reason for this is that rank-0 units are deemed unimportant by our ranking process. Adding new incoming connections enhances their capacity and gives them a higher chance to be vital in the future. Also, new connections let them reuse knowledge of higher-rank units, promoting knowledge transfer. Hence, we randomly create new connections from all units to rank-0 units with zero weights.

\subsection{Unit States and Forming Stable Connections}
Besides ranks, we introduce unit states. Rank-0 units are in the ``Idle'' state because they are not allocated for the current episode. Idle units are disconnected from loss, so they do not receive training (see \textbf{Figure}~\ref{fig:main_figure}-b). We observe that reinitializing these idle units before each episode leads to better performance. A similar observation is done in a recent work \cite{Continual_Backprop} suggesting SGD loses plasticity as training progresses and reinitializing unimportant units leads to improved performance. Rank-1 units are in ``Training'' as they were previously idle, but now they are being trained.

We have two additional states: ``Fine-tuning'' and ``Frozen''. These states depend not only on the unit's rank but also on the unit's position and the capacity of the STM. Recall that we divide the DNN into two modules: $G(\cdot)$ (before replay) and $F(\cdot)$ (after replay). Also, recall that our STM holds activations from $m$ previous episodes. If a unit has rank $r$ where $r \geq 2$ or $r \leq m+1$ and it is in $F(\cdot)$, which receives replays, we call it a fine-tuning unit. The intuition is that such units were initially allocated for a past episode, but they can still be fine-tuned on new episode classes without forgetting because the classes they encountered earlier are still in the STM that remembers $m$ previous episodes.  

If a unit has rank $r > m +1$ and is in $F(\cdot)$, it risks forgetting because the STM has a temporal capacity of $m$ and will purge some classes that have been seen by the unit. So, we must stop training such units and all their ancestors to retain knowledge about classes that will be purged from the STM.

{\centering
    \emph{\textbf{Freezing Rule: } Freeze all incoming connections to rank $r > m +1$ units in $F(\cdot)$.}
\par}
This rule, combined with the \emph{Path Property}, handles all ancestors of a unit because the \emph{Path Property} suggests that all ancestors of a unit have either the same rank or a higher rank. Consequently, they will either already be frozen or will be frozen simultaneously.

Finally, we freeze incoming connections to units with rank $r > 1$ if they are in $G(\cdot)$ --- this is the same \emph{Freezing Rule}, but we set $m = 0$ since $G(\cdot)$ does not replay. See \textbf{Figure}~\ref{fig:main_figure}-c for a diagram of states and ranks. Note that we are not freezing all the units that generate activations we store for replay. So, dimensions of activations corresponding to rank $0$ and $1$ may arbitrarily change with new episodes and be unreliable for replay. However, this is not an issue because their activations will not be propagated to Fine-tuning units with $r > 1$  that rely on replay to remember old knowledge (recall \emph{Path Property}). So, we can safely replay partially drifted representations. This lets SHARP continuously learn to extract features without relying on pretrained and frozen layers.

\subsection{Supervised Contrastive Loss}
Typically, DNNs use cross-entropy loss and train a classification layer. However, this can lead to problems in CL, as the network's predictions are very sensitive to the weights of the classification layer. For example, \cite{last_layer} suggests several strategies to address this issue, such as normalizing weights or modifying bias terms. Also, it is common practice in CL to partition the output unit into multiple classification heads or mask out certain output units to prevent large drifts in decision boundaries \cite{scenarios}. We take a different approach and remove the classification layer altogether and instead learn representations that are clustered based on classes through Supervised Contrastive Learning \cite{SUPCON}.

During the forward pass, we start with a batch of $n$ input samples. Then, we feed these samples through $G(\cdot)$ and get $n$ hidden activations. Next, we sample additional $n$ activations from STM and have a total of $2n$, which is fed to $F(\cdot)$ that outputs $2n$ representations (see \textbf{Figure}~\ref{fig:main_figure}-b). Let us denote indices of these $2n$ representations $\left\{z_1, \dots z_{2n} \right\}$ with set $I$. Let $i \in I$ be the index of an arbitrary sample in $I$. We define two sets: (1) $A(i) \equiv I \setminus \left\{ i \right\}$, \textit{\textbf{A}ll except $i$}, (2) $P(i) \equiv \left\{ p \in A(i) : y_p = y_i \right\}$, \textit{\textbf{P}ositive with respect to $i$}. Then, we compute the following loss function:

\begin{equation}
    \mathcal{L} = \sum_{i \in I} \frac{-1}{\left| P(i) \right|} \sum_{p \in P(i)}\log \frac{\exp(< \hat{z}_i \cdot \hat{z}_p > / \lambda)}{\sum_{a \in A(i)} \exp(< \hat{z}_i \cdot \hat{z}_a > / \lambda)}
\end{equation}

Here we modify $z_i$ by setting all rank-0 unit activations to 0 and normalize the resulting vector to have a unit L2 norm to obtain $\hat{z}_i$. Also, $< \cdot >$ is the dot product, and $\lambda$ is a temperature parameter -- usually between $0.05$ to $0.2$. This loss clusters examples of the same class in embedding space while pushing apart clusters of samples of different classes. Consequently, our network learns representations but cannot make predictions. We rely on a k-NN classifier for predictions, as explained next.

\subsection{Long Term Memory and Predictions}
After each episode, we randomly select a subset of activations from the STM and transfer them to the Long Term Memory (LTM) using $F(\cdot)$. At this point, the STM has activations for the current episode as well as the previous $m$ episodes. When generating LTM representations for the classes of episode $e-i$, we discard dimensions from units with rank $< m+i$ since these units were not trained on the classes of episode $e-i$ and are essentially just noise. It is also unsafe to use these dimensions for predictions since they will be updated in future episodes. For example, rank-0 units are disconnected from the loss during learning for episode $e$, so they can be discarded when predicting episode $e$ classes. Additionally, they will be updated in future episodes, so it's not safe to store them in LTM since they will become outdated. This same logic applies recursively to other ranks and episodes. 

Note that LTM representations occupy negligible space compared to the STM activations because we select a random subset to store,  representations are compressed due to decreasing layer widths, and we only store dimensions corresponding to certain ranks. Finally, SHARP maps a given test example to a class. We process sample $x_i$, yielding $z_i = F(G(x_i))$. Then we perform  k-NN classification using $z_i$ and LTM representations, employing a normalized L2 distance:
     $\left\| \phi(c_j * m_j) - \phi(z_i * m_j) \right\|_2^2$ Here, $c_j$ is a representation in LTM, $\phi$ is L2 normalization and $m_j$ is a mask that selects stored dimensions.

\section{Experimental Results}
This section compares SHARP with state-of-the-art CL methods on five CIL sequences. The first three sequences consist of five learning episodes, derived from the MNIST, FMNIST, and CIFAR10 datasets. Each episode includes two adjacent classes. The fourth is based on EMNIST and includes 26 English letters. Lowercase and uppercase letters are mapped to the same class. Lastly, the fifth is based on the CIFAR100 dataset, where one learning episode has 10 consecutive classes, resulting in 10 episodes. SHARP's STM only holds activations from the previous episode (i.e., $m = 1$). We measure replay memory sizes in megabytes for all baselines, as detailed in Supplementary Material.
 
For MNIST, FMNIST, and EMNIST, our network has two convolutional layers (16 filters) and one hidden linear layer with 500 units, followed by an output layer. In the case of CIFAR10 and CIFAR100, we use seven convolutional layers and a single hidden linear layer with 1024 units, leading to an output layer. This architecture is based on VGG-16, but we reduced its depth and width since our datasets are smaller. In all baseline models, the width of the output layer is equal to the number of classes. However, since SHARP learns representations, its output layer maintains the same width as the preceding layer. This does not imply that SHARP has more parameters. On the contrary, SHARP uses sparse networks and prunes 60\% of connections (fixed hyperparameter), resulting in fewer parameters overall. For further information, please refer to the Supplementary Material.

\begin{figure}
  \centering
  \includegraphics[width=\textwidth]{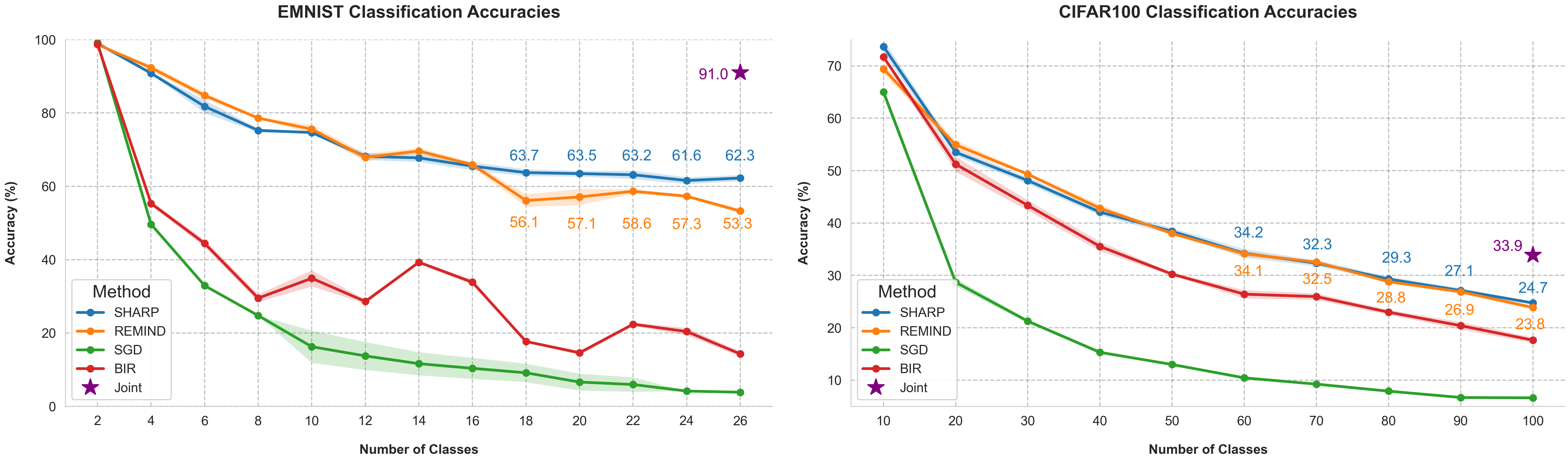}
  \caption{Accuracies on seen classes for EMNIST and CIFAR100. We do not consider performances in pretraining classes. Results are across three random seeds (with the same pretrained weights). EMNIST memory budget: 0.0392 MB. CIFAR100 memory budget: 4.096 MB. BIR does not have explicit memory but has additional variational autoencoder parameters for generating old activations.}
  \label{fig:representation_replay_results}
  \vskip -0.2in
\end{figure}

\subsection{Comparison with Activation Replay Methods}
We compare SHARP with two activation replay methods, REMIND and Brain-inspired Replay (BIR). As mentioned  in the Related Work section, these methods assume that layers not subjected to replay have undergone pretraining and frozen. So, we pretrain them on MNIST and then sequentially teach them EMNIST and CIFAR10, followed by sequential learning of CIFAR100. Note that SHARP does not require such pretraining and freezing, as it can continually learn representations even in earlier layers. However, this presents SHARP with a significantly more challenging scenario. Therefore, we also pretrained the module $G(\cdot)$ for SHARP to have  a fair comparison. Also, we set the ranks of units in $G(\cdot)$ to the maximum rank possible (i.e., $E$) to ensure the ranking logic works. Finally, we omitted the pruning step for $G(\cdot)$ since it is unnecessary to disentangle frozen and pretrained units.

SHARP's and REMIND's replay cut-off is after the first two layers for EMNIST and after the first four layers for CIFAR100. We pretrain/freeze all convolutional layers for BIR as suggested in \cite{gido2020}. REMIND is proposed for online learning --- single epoch and batch size of 1. Therefore, we adapted it for a batch setting and multiple epochs following \cite{remind}. See Supplementary Material for details.

\textbf{Figure}~\ref{fig:representation_replay_results} displays the results. Firstly, we observe that BIR performs worse than both REMIND and SHARP. We suspect that this is because BIR relies on generated representations. As a result, the performance of BIR heavily relies on the quality and diversity of the generated activations. It can be challenging to ensure that these activations are of high quality, particularly in a continual setting. In contrast, REMIND and SHARP replay actual activations, which are more dependable and robust. We believe this is why REMIND and SHARP can achieve better performance than BIR.

Secondly, we notice that REMIND matches or slightly outperforms SHARP when the number of seen classes is small. However, as we see more classes, SHARP surpasses REMIND. This is expected because as we progress in learning, REMIND needs to uniformly replay all past classes and shred fixed memory for more classes. Thus, REMIND's job becomes harder as we see more classes, while SHARP only stores and replays the classes from the previous episode. This makes SHARP's replay agnostic to the number of seen classes and more robust for longer sequences. Finally, it is typical for CL methods to exhibit lower performance than the model that learns all classes jointly, which represents the upper bound for CL. This motivates future research to close the performance gap.

\subsection{Comparison with Raw Replay Methods}
We compare SHARP against state-of-the-art raw replay methods. In contrast to REMIND and BIR, SHARP and raw replay methods can continually extract features, so we did not perform pretraining. Raw replay baselines have an unfair advantage since they replay images rather than activations and all past classes instead of just few recent ones. To ensure a fair comparison, we used a fixed memory budget, so both SHARP and the baselines had an equal footing in terms of the megabytes used for storing replay samples. See the Supplementary Material for more details and additional results.

\textbf{Table}~\ref{tab:classification_accuracies} shows the accuracy results for SHARP and five baseline methods, along with Joint training (upper bound) and standard SGD (lower bound), across all classes after all episodes. We observe that SHARP significantly outperforms all baselines on all datasets except for iCaRL on CIFAR10.  To our knowledge, this is the first time a CL approach has surpassed replay-based methods in a CIL context without requiring the replay of all previously encountered classes. We hope this accomplishment will inspire future research to focus on neuro-inspired replay instead of attempting to approximate static data distributions by blending new data with selected samples from all past classes.  

Besides SHARP, iCaRL also shows strong performance. Interestingly, iCaRL uses a nearest-mean-of-exemplars classifier, which is similar to the k-NN mechanism used in our LTM. This similarity motivates us to closely examine iCaRL for potential improvements that could be applied to SHARP in future research. For example, instead of randomly selecting samples as SHARP does, iCaRL prioritizes selecting samples that cause the average penultimate layer activation over all stored class samples to best approximate the average activation over all seen examples of a class. This ensures that iCaRL has the best class mean vector in the classifier. A similar technique could be applied to SHARP. Furthermore, this may explain why iCaRL outperforms SHARP on CIFAR10 since random selection might not provide SHARP with optimal representations for the k-NN classifier, particularly in datasets like CIFAR10 where examples within the same class can differ significantly (e.g., two handwritten ``A''s might appear quite similar, but two cat images can vary greatly).

\begin{table}[ht]
\centering
\renewcommand{\arraystretch}{1.2}
\caption{Average accuracy across classes after all episodes. MNIST/FMNIST/EMNIST memory: 0.0392 MB (50 images), CIFAR10 memory: 1.6384 MB ($\approx534$ images), CIFAR100 memory: 4.096 MB ($\approx1334$ images). See Supplementary Material for results with other budgets.}
\begin{tabularx}{0.9\textwidth}{Y|Y|Y|Y|Y|Y}
\hline \hline
\textbf{Method}    & \textbf{MNIST} & \textbf{FMNIST} & \textbf{EMNIST} & \textbf{CIFAR10} & \textbf{CIFAR100} \\ \hline \hline
Joint               & 98.0 ($\pm 0.0$)          & 86.6 ($\pm 1.0$)             & 91.0 ($\pm 0.3$)            & 70.8 ($\pm 0.4$)             & 33.9 ($\pm 0.8$)               \\ \hline
\textbf{SHARP}               & \textbf{88.0} ($\pm 1.3$)            & \textbf{72.5} ($\pm 2.5$)             & \textbf{55.9} ($\pm 3.1$)             &  46.3 ($\pm 2.7$)              & \textbf{23.1} ($\pm 0.2$)              \\
iCaRL\cite{icarl}              & 85.1 ($\pm 0.5$)            & 70.7 ($\pm 0.7$)             & 55.1 ($\pm 0.6$)             & \textbf{53.0} ($\pm 0.8$)              & 21.4 ($\pm 0.5$)               \\
DER \cite{DER}                & 82.7 ($\pm 2.7$)            & 72.2 ($\pm 0.4$)            & 44.6 ($\pm 4.1$)             & 37.6 ($\pm 0.9$)           & 16.6 ($\pm 0.3$)               \\
ER \cite{EReplay}                 & 69.1 ($\pm 2.7$)            & 66.8 ($\pm 1.6$)             & 29.9 ($\pm 0.9$)             & 32.8 ($\pm 1.0$)             & 11.4 ($\pm 0.1$)              \\
FDR \cite{fdr}               & 76.5 ($\pm 1.1$)            & 65.0 ($\pm 4.2$)             & 32.5 ($\pm 2.1$)             &      30.1 ($\pm 4.3$)        & 8.7 ($\pm 0.5$)                \\
A-GEM \cite{agem}               & 30.0 ($\pm 5.5$)            & 40.9 ($\pm 1.9$)             & 9.5 ($\pm 0.3$)             & 19.5 ($\pm 0.7$)             & 6.5 ($\pm 0.2$)               \\ \hline
SGD               & 19.8 ($\pm 0.1$)            & 16.6 ($\pm 4.7$)            & 3.9 ($\pm 0.0$)             & 16.0 ($\pm 4.2$)              & 6.6 ($\pm 0.1$)               \\

\hline \hline
\end{tabularx}
\label{tab:classification_accuracies}
\end{table}

\subsection{Does SHARP Need Episode Boundaries?}
In our earlier experiments, we demonstrated the worst-case scenario of CIL, where each class appears just once, and all episodes contain distinct classes. However, some of SHARP's steps, like assigning specific units to some episodes or freezing certain connections, could be perceived as taking advantage of this highly organized and non-overlapping arrangement. In the real world, the difference between episodes may become blurry, and they may have overlapping classes.

To answer this, we introduce a straightforward way of generating noisy CIL sequences and test SHARP's behavior on them. We create these noisy sequences by initializing two sets, $N$ and $S$, set of novel and previously seen classes, respectively. In the beginning, all classes are in $N$, and $S$ is empty. We then have two hyperparameters: $c$, the number of classes per episode, and $\eta$, the probability of substituting a class from the previous episode with a new one – we replace it with a seen class with a probability of $1 - \eta$. In this case, standard joint training occurs when $\eta = 0$ and $c =$ the total number of classes. Conversely, if $\eta = 1$, we have a typical CIL scenario with non-overlapping episodes. In other words, increasing $\eta$ shifts the setting from i.i.d. to CIL. 

\begin{figure}
  \centering
  \includegraphics[width=0.90\textwidth]{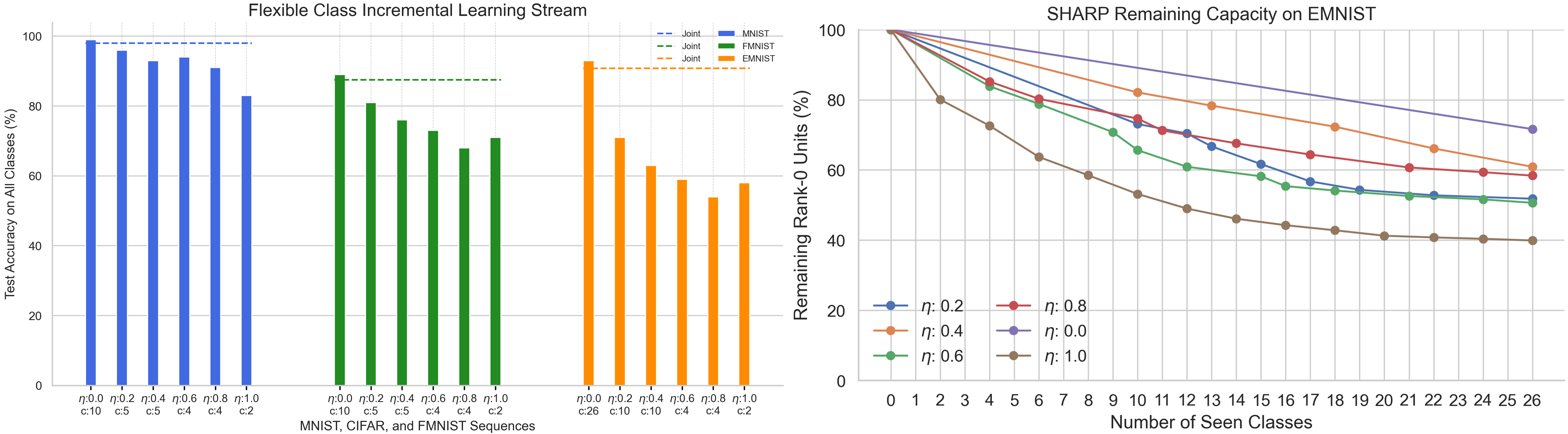}
  \caption{L) Accuracy on all classes.  R) Remaining rank-0 units as the number of classes increases.}
  \label{fig:remanining_rank0_ratio}
  \vskip -0.15in
\end{figure}

\textbf{Figure}~\ref{fig:remanining_rank0_ratio}-left shows the accuracy after all episodes for various sequences created by varying $\eta$ and $c$. First, a smaller value of $\eta$ generally leads to better performance. This is intuitive because a smaller value of $\eta$ means that SHARP is more likely to see the same class multiple times, allowing SHARP to learn the class more thoroughly and improve its accuracy. 

Another critical question is whether encountering the same classes multiple times will trigger recruiting more rank-1 units and lead to quickly reducing rank-0 units. If this happens, SHARP will run out of capacity since  rank-0 units are reserved for the future and indicate the model’s remaining learning capacity. \textbf{Figure}~\ref{fig:remanining_rank0_ratio}-right illustrates changes in rank-0 units. Here, we see that as SHARP becomes more competent in detecting classes, it needs to allocate fewer and fewer units to learn more classes. In other words, SHARP relies on past knowledge to quickly learn new classes without reducing learning capacity. Moreover, if two consecutive episodes are similar (e.g., only one new class), it employs few rank-0 units to learn, which means SHARP has implicit novelty detection capabilities. Overall, although some of the algorithmic steps of SHARP rely on episode boundaries, it is robust in scenarios where episodes become blurry and may include overlapping classes.

\section{Conclusion and Future Work}
SHARP is a novel neuro-inspired  replay approach for continual learning. To the best of our knowledge, it is unique in that it can avoid forgetting by only replaying recently encountered classes. Furthermore, unlike other activation replay methods, it does not require a pretrained feature extractor; instead, it can learn to extract features continually. We demonstrated that SHARP outperforms state-of-the-art methods on several benchmark datasets. Also, we have shown that SHARP is flexible and handles scenarios where episodes are not highly structured and may contain overlapping classes.

SHARP's replay is similar to NREM (Non-rapid eye movement) sleep and awake replay in the brain. However, it does not include REM sleep, an essential part of memory consolidation in the brain. During REM sleep, the brain organizes long-term memories without external input to prevent interference between memories and improve generalization. In future work, we aim to incorporate a metric learning module into our LTM to enhance the generalization of LTM representations without external input. This module will continuously update the distance function used for classification based on existing memories, similar to memory reorganization during REM sleep. 
\newpage

\bibliography{main}
\bibliographystyle{bib}

\newpage

\section*{Supplementary Material}

\setcounter{section}{0} 

\renewcommand{\thesection}{\Alph{section}} 

\section{Greedy Algorithm Optimality Proof}
Recall that we compute the activation at each layer on some episode examples as follows:

\begin{equation}
    A_l =  \sum_{x \sim D_e} a_l(x) =  \sum_{x \sim D_e} \sum_{u_i^l \in U^{l}}   a_{u_i^l}(x)
\end{equation}

Here, $a_{u_i^l}(x)$ is the activation of unit $i$ at layer $l$ for sample $x$ and $a_l(x)$ is the total activation of layer $l$ for sample $x$. Our ranking system relies on solving the following constrained optimization problem:

\begin{equation}
     \underset{P^l_{1} \subseteq  U^{l}_{r < 2}}{\min} |P^l_{1}| \; \; \text{subject to}\; \; \sum_{x \sim D_e}   \sum_{u_i^l \in P^l_{1}} a_{u_i^l}(x) \geq \tau  A_l  - \sum_{u_i^l \in U^{l}_{r \geq 2}} a_{u_i^l}(x)   
\end{equation}

 where $0 \leq \tau \leq 1$ and $P^l_{1}$ is set of units we promote to rank-1. Notice that the right side of the inequality does not contain the set in the minimization objective. So, we replace it with a constant: 

 \begin{equation}
     \underset{P^l_{1} \subseteq  U^{l}_{r < 2}}{\min} |P^l_{1}| \; \; \text{subject to}\; \; \sum_{x \sim D_e}   \sum_{u_i^l \in P^l_{1}} a_{u_i^l}(x) \geq T 
\end{equation}
Here, we have a set of units $U^{l}_{r < 2}$ with a rank less than 2 that we can select. Unit $i$ in this set has total activation of $\sum_{x \sim D_e}  a_{u_i^l}(x)$. Our goal is to pick a minimum number of units in $U^{l}_{r < 2}$ to achieve or exceed target $T$. We propose the following simple greedy algorithm:
 \begin{itemize}
     \item Sort units in $U^{l}_{r < 2}$ based on their activations $\sum_{x \sim D_e}  a_{u_i^l}(x)$ in descending order.
     \item Pick the unit with the largest total activation and add to $P^l_{1}$ until we reach or exceed target $T$.
 \end{itemize}

\textbf{Proof: }We will show that the greedy algorithm finds a solution with a sum of activations at least $T$, using fewer or the same number of units compared to any other optimal algorithm. To prove this, we compare the partial sums of activations for both algorithms at each step. Let $A_k$ be the sum of activations for the first $k$ units (i.e., $A_k = \sum_{i =1}^k a_i$) chosen by the greedy algorithm, and $B_k$ be the sum of activations for the first $k$ units (i.e., $B_k = \sum_{i =1}^k b_i$) chosen by the optimal algorithm.

We use induction to prove that $A_k \geq B_k$ for all $k$. The base case is trivial, as $A_0 = B_0 = 0$. For the inductive step, suppose that $A_{k-1} \geq B_{k-1}$. Then, since the greedy algorithm selects the unit with the highest activation at each step, we have $a_k \geq b_k$. Therefore, $A_k = A_{k-1} + a_k \geq B_{k-1} + b_k = B_k$. Thus, by induction, we have $A_k \geq B_k$ for all $k$. The greedy algorithm is guaranteed to reach the target sum $T$ by selecting fewer or the same number of units as the optimal algorithm.

\newpage

\section{SHARP Pseudocode}

\begin{algorithm}
\caption{SHARP algorithm for a single Episode}
\begin{algorithmic}
\setstretch{1.5}
\Require{$D_e$, $G$, $F$, STM, LTM}
\Comment{Require Episode Dataset, $G$ module, $F$ module, STM, LTM}
\For{every phase $p$ in the Episode}
    \State $\tau_p \gets \max(\tau_{min}, \frac{1}{2} \left ( 1 + \cos \left ( \frac{p+1}{k} \times \pi \right ) \right ))$
    \Comment{Determine $\tau_p$ using cosine annealing schedule}
    \State $P^l_{1} \gets \text{Selection}(D_e, G, F, \tau_p)$
    \Comment{Select rank-1 units using activations}
    \State $G, F \gets \text{Drop}(G, F)$
    \Comment{Drop connections from rank-0 to rank-1 Unit}
    \State $G, F \gets \text{Grow}(G, F)$
    \Comment{Grow an incoming connection to rank-0 for every dropped one}
    \State $G, F \gets \text{Train}(G, F, D_e, \text{STM})$
    \Comment{Train the network for a fixed number of epochs}
\EndFor

\State \# End of an episode 
\State $G, F \gets \text{Promote}(G, F)$
\Comment{Increment all ranks greater than 0 by one}
\State $G, F \gets \text{Freeze}(G, F)$
\Comment{Freeze incoming connections based on the freezing rule}
\State $\text{STM} \gets \text{Update\_STM}(G, \text{STM}, D_e)$
\Comment{Sample examples from $D_e$ and push activations to STM}
\State $LTM \gets \text{Update\_LTM}(F, \text{STM})$
\Comment{Create few LTM representations for classes in STM}
\State $\text{STM} \gets \text{Purge\_STM}(\text{STM})$
\Comment{Remove oldest classes to make space}
\State $G, F \gets \text{Reinit\_Rank0}(G, F)$
\Comment{Reinitialize incoming connections of rank-0 Units}
\State \# SHARP is ready to process next episode
\end{algorithmic}
\end{algorithm}

\section{Experimental Details}

\subsection{Architectural Details}
In our experiments, we employed two different architectures -- one for MNIST/FashionMNIST/EMNIST and another for CIFAR10/CIFAR100. The complete specifications for each architecture are provided in \textbf{Listing}~\ref{lst:mnist_architecture} and \textbf{Listing}~\ref{lst:cifar_architecture}.

The architectures for Brain-inspired Replay (BIR) differ slightly due to its generative nature. BIR includes additional generative weights, which consist of reversed versions of the linear layers and a few sets of extra weights. Moreover, for CIFAR10/100, BIR employs two linear layers with 2000 units each, instead of a single linear layer with 1024 units used in other baselines. We made this modification to ensure that BIR has a reasonable number of learnable parameters since it freezes all the convolutional layers after pretraining \cite{gido2020}. As a result of its generative components and the additional parameters introduced for CIFAR10/100, BIR has more parameters compared to other replay methods, including SHARP. However, we believe this is fair since BIR does not store activations or input samples explicitly. Instead, it relies on these additional weights to generate them. Therefore, its memory budget is allocated towards additional learnable parameters. For more details, please refer to the Brain-inspired Replay paper \cite{gido2020}.

For SHARP and REMIND \cite{remind}, we partitioned the architecture into layers that receive replay (i.e., $F(\cdot)$) and those that do not (i.e., $G(\cdot)$). As seen in \textbf{Listing}~\ref{lst:mnist_architecture} and \textbf{Listing}~\ref{lst:cifar_architecture}, SHARP and REMIND split the networks after the second and fourth convolutional layers. In contrast, BIR only replays through the linear layers, while all the convolutional layers are pretrained and frozen, as suggested in the original implementation.

Choosing where to split the architecture is crucial as the replay cut-off directly impacts the amount of memory required to store the activations. If the cut-off is set too low, the stored activations will occupy a large amount of space. Conversely, if it is set too high, only a few layers will benefit from replay, leading to poor performance. Therefore, it is essential to strike a balance between storing smaller-sized activations and replaying through more layers. In addition to this trade-off, SHARP offers flexibility in terms of how the network is split into two. It is even possible to start the replay from the input, which provides a raw replay version of SHARP without any modification. However, we did not run any experiments with this version since the core focus of our work is to avoid replaying raw examples.

\subsection{Linear Quantization and Memory Budget}
The Short Term Memory (STM) used by SHARP employs linear quantization to efficiently store floating point activations using a single byte. Linear quantization involves mapping continuous numerical values to a fixed number of discrete levels, reducing precision but increasing storage efficiency. To implement this, the tensor being stored is first min-max scaled and then multiplied by a quantization range of $2^8 - 1$ (with 8 representing the number of bits in a byte). The resulting normalized tensor is rounded to the nearest integer and represented using an unsigned 8-bit integer. During dequantization, the same steps are performed in reverse order. While more advanced quantization techniques exist to reduce reconstruction error and improve compression ratios, we did not focus on optimizing this process.

Now, let us compute the memory usage of a single example in megabytes for our experiments. For MNIST/FMNIST/EMNIST, we store activations after the second convolutional layer. In this case, we have 16 feature maps of size $7 \times 7$. Each entry can be stored using a single byte. Therefore, the total memory usage for activations of a single example can be calculated as follows:

\begin{equation*}
16 \times 7 \times 7\ byte = 0.000784\ MB
\end{equation*}

If we were to store raw examples instead of the activations, we would need $28 \times 28 \times 1$ bytes, which is also equal to 0.000784 MB. Therefore, in these experiments, both raw examples and activations occupy the same amount of memory.

For CIFAR10 and CIFAR100, we store activations after the fourth convolutional layer. In this case, we have 64 feature maps, each with a size of $8 \times 8$. Therefore, the memory usage for a single example when storing activations is:

\begin{equation*}
64 \times 8 \times 8  = 0.004096\ MB
\end{equation*}

However, if we were to store raw examples, they would require $32 \times 32 \times 3$ bytes, which is equivalent to 0.003072 MB. Therefore, it is more efficient to store raw examples in this case. It is important to note that these calculations are specific to the architecture and replay cut-off. Thus, if memory efficiency is a concern, SHARP's replay cut-off or architecture can be adjusted accordingly.

SHARP does not have an advantage in terms of the number of entries stored in memory in any of our experiments. However, the true strength of SHARP lies in its ability to store a limited number of recently seen classes. This allows it to store significantly more replay examples per class, providing an effective solution for mitigating catastrophic forgetting.

Finally, some of the replay baselines incorporate extra information alongside the stored samples or activations. For instance, DER and FDR store the network outputs for each replayed sample, while SHARP maintains few long-term representations per class. To maintain simplicity in comparisons, we do not consider these additional memory usages as part of the fixed memory budget. It's worth noting that these supplementary usages are negligible compared to the stored samples or activations.

\subsection{Sparsification Details}
Another  detail of SHARP is the utilization of sparse connectivity. Initially, architectures are made sparse through unstructured random pruning on a per-layer basis. This means that each connection within a layer has an equal probability of being removed, resulting in consistent density across all layers. We maintained a sparsity level of 60\% or equivalently, a density of 40\% in all of our experiments. To keep things simple, we did not optimize this choice for each experiment individually. Instead, we based our decision on coarse hyperparameter finetuning on CIFAR100 and applied the 60\% sparsity level consistently across all datasets. Please note that all other baselines are densely connected and have significantly more learnable parameters than SHARP.

In convolutional layers, a ``unit'' refers to a 3D convolution filter, while a ``connection'' represents a 2D kernel rather than an individual weight. Consequently, when we take actions like dropping or freezing a connection, we consider the entire 2D kernel that represents the connection between two units. This formulation significantly reduces the number of floating point operations (FLOPs) by eliminating operations performed on the majority of 2D feature maps.

It is essential to emphasize that the first layer in convolutional architectures should not be pruned. The reason behind this is that the first layer receives a three-channel image as input, and its units are connected to the input through only three connections. If we were to prune the first layer, it would likely result in the creation of numerous "dead units" that have no inputs at the initial layer. For example, at a density of 0.2, the probability of units in the first layer becoming dead would be $(1 - 0.2)^3 = 0.512$. To avoid this issue, it is advisable to maintain the density in the first layer. Importantly, this change has minimal impact on the overall number of parameters.

\begin{center}
\begin{lstlisting}[basicstyle=\small, caption=MNIST/FMNIST/EMNIST architecture, label=lst:mnist_architecture]
[G1] conv2d(in=3, out=16, kernel=3, stride=1)
[G2] nn.ReLU()
[G3] nn.MaxPool2d(kernel=2, stride=2)
[G3] conv2d(in=16, out=16, kernel=3, stride=1)
[G4] nn.ReLU()
[G5] nn.MaxPool2d(kernel=2, stride=2)
---------------------- Second Module Starts ----------------------
[F1] linear(in=16 * 7 * 7, out=500)
[F2] nn.ReLU()
[F3] linear(in=500, out=num_classes (500 if SHARP))
[F4] nn.Sigmoid() # If SHARP nn.ReLU()
\end{lstlisting}
\end{center}

\begin{center}
\begin{lstlisting}[basicstyle=\small, caption=CIFAR10/CIFAR100 architecture, 
label=lst:cifar_architecture]]
[G1] conv2d(in=3, out=64, kernel=3, stride=1)
[G2] nn.ReLU()
[G3] conv2d(in=64, out=64, kernel=3, stride=1)
[G4] nn.ReLU()
[G5] nn.MaxPool2d(kernel=2, stride=2)
[G6] conv2d(in=64, out=64, kernel=3, stride=1)
[G7] nn.ReLU()
[G8] conv2d(in=64, out=64, kernel=3, stride=1)
[G9] nn.ReLU()
[G10] nn.MaxPool2d(kernel=2, stride=2)
---------------------- Second Module Starts ----------------------
[F1] conv2d(in=64, out=128, kernel=3, stride=1)
[F2] nn.ReLU()
[F3] conv2d(in=128, out=128, kernel=3, stride=1)
[F4] nn.ReLU()
[F5] conv2d(in=128, out=128, kernel=3, stride=1)
[F6] nn.ReLU()
[F7] nn.MaxPool2d(kernel=2, stride=2)
[F8] linear(in=128 * 4 * 4, out=1024)
[F9] nn.ReLU()
[F10] linear(in=1024, out=num_classes (1024 if SHARP))
[F11] nn.Sigmoid() // If SHARP nn.ReLU()
\end{lstlisting}
\end{center}

\subsection{SHARP's Hyperparameters}
While SHARP does have a wide range of hyperparameters, our observations indicate that the majority of them do not have a significant impact on its performance. In fact, default values tend to yield satisfactory outcomes across multiple experiments. In this subsection, we will provide a comprehensive overview of all the hyperparameters and offer recommended default values for each.

\begin{itemize}
    \item For STM's temporal window size, we set $m=1$ for all experiments, but any value can be used without modifying the algorithm.
    
    \item The number of epochs per phase $\epsilon$ is typically set to 3 or 5 for good performance, but more complex datasets may require more epochs.
    
    \item The minimum value for the cosine annealing schedule, $\tau_{min}$, determines how small $\tau$ can be reduced. If $\tau_{min}$ is small, SHARP will recruit fewer rank-0 units, which can hurt performance but reserve more capacity for the future. For longer sequences, a smaller $\tau_{min}$ is suggested, while a larger $\tau_{min}$ is better for shorter sequences.
    
    \item The number of phases $\pi$ is determined by the complexity of the dataset, and $\pi \times \epsilon$ gives the total number of training epochs per an episode.

    \item The parameter $k$ in the cosine annealing schedule determines the step sizes for decreasing $\tau$. We set $k=30$ for all experiments.
    
    \item The number of neighbors to use in LTM's k-NN is not very sensitive, and 5 can be a good default hyperparameter.
    
    \item We used 25 (MNIST/FMNIST/EMNIST) or 50 (CIFAR10/100) LTM representations per class in our experiments, but this value cannot exceed the number of activations STM stores per class, as LTM representations are created using STM entries.

    \item Supervised Contrastive Learning Loss temperature term $\lambda$. This is not exclusive to SHARP, but it is typically set between 0.05 to 0.2 in Supervised Contrastive Learning.
\end{itemize}

\textbf{Table}~\ref{tab:sharp_hp_table}, we provide a comprehensive list of all the hyperparameters that are unique to SHARP. It should be noted that we performed systematic finetuning for the $\tau_{min}$ hyperparameter, tried $\lambda = 0.05, 0.1, 0.2$ and tried 5 and 25 for LTM's k-NN neighbors, while the remaining hyperparameters were selected based on educated guesses. 
The following subsections provide standard hyperparameters, including learning rate, batch size, and optimizer details, for both SHARP and all other baselines.

\begin{table}[ht]
\newcolumntype{C}[1]{>{\centering\arraybackslash}p{#1}}
\centering
\small
\renewcommand{\arraystretch}{1.25}
\caption{SHARP specific hyperparameters used in main experiments. Interestingly setting LTM's number of neighbors to 25 gave better performance for FMNIST, so we kept it to 25. We only tried 5 or 25 during hyperparameter finetuning.}
\begin{tabularx}{\textwidth}{C{2cm}|Y|Y|Y|Y|Y|C{2.2cm}|C{2cm}|Y}
\hline \hline
 Experiments   &  $m$ & $\epsilon$ & $\tau_{min}$ & $\pi$ & $k$  & LTM's neighbors  & LTM per class & $\lambda$ \\ 
  \cline{1-9} \multicolumn{9}{c}{Experiments from main paper's Section 4.2} \\ \cline{1-9}
MNIST & 1 &  3 & 0.90 & 10 & 30 & 5 & 25 & 0.1 \\
FMNIST & 1&  3 & 0.75 & 12 & 30  & 25 & 25 & 0.2 \\
EMNIST & 1 &  3 & 0.60 & 15 & 30  & 5 & 25 & 0.1 \\
CIFAR10 & 1&  5 & 0.70 & 15 & 30 & 5 & 50 & 0.2 \\
CIFAR100 & 1 & 5 & 0.70 & 15 & 30  & 5 & 50 & 0.05 \\
 \cline{1-9} \multicolumn{9}{c}{Below are experiments with pretraining (main paper's Section 4.1)} \\ \cline{1-9}
EMNIST& 1 &  3 & 0.40 & 15 & 30  & 5 & 25 & 0.05 \\
CIFAR100 & 1 &  5 & 0.70 & 15 & 30  & 5 & 50 & 0.05 \\
\hline \hline
\end{tabularx}
\label{tab:sharp_hp_table}
\end{table}

\subsection{Details for Comparison with Activation Replay Methods}
In Section 4.1 of the main paper, we compared SHARP with REMIND and BIR. As REMIND and BIR require pretraining, we first pretrained on MNIST before sequentially teaching EMNIST, and on CIFAR10 before sequentially teaching CIFAR100. To ensure a fair comparison, we repeated the pretraining for SHARP as well. We only loaded pretrained weights to layers that did not receive replay and reinitialized the plastic layers for all approaches.

It should be noted that SHARP uses supervised contrastive learning loss for pretraining, as it does not have a classification layer. Additionally, we did not perform pruning to module $G(\cdot)$ for SHARP since it is not necessary to disentangle pretrained layers. Finally, we set the units in $G(\cdot)$ to the maximum rank possible, meaning that those units are frozen and do not drop any connections.

\textbf{REMIND details:} REMIND is a neuro-inspired approach to replay compressed representations. It was originally designed for online learning, processing one example at a time and performing a single pass over datasets instead of multiple epochs. However, it is possible to implement a batch version of REMIND and perform multiple epochs. In fact, the original paper shows that the batch version leads to better performance than online learning. As both SHARP and BIR learn with batches, we implemented the batch version of REMIND for a fair comparison.

Batch REMIND first passes $n$ input samples through its pretrained and frozen layers. Then, it samples $n$ hidden layer activations from memory, which are uniformly distributed across all previously seen classes. To store compressed representations, REMIND utilizes Product Quantization (PQ), a machine learning technique that compresses high-dimensional vectors by dividing them into subvectors and quantizing each subvector separately. This is typically done using a clustering algorithm such as k-means, which groups similar subvectors together to reduce the overall dimensionality of the data. PQ has two hyperparameters: the size of subvectors and the number of codebooks. We set the number of codebooks to 256, which allows each codebook to be represented with a single byte ($2^8$). Additionally, we experimented with several subvector sizes to find the optimal trade-off between reconstruction error and compression ratio.

In the original implementation of REMIND, data augmentation was used for input samples, and a variant of manifold mixup was applied to the quantized tensors to augment replay activations. However, we did not employ any form of data augmentation for our baselines. To ensure a fair comparison, we excluded both the data augmentation and manifold mixup steps from REMIND in our evaluation.

For our experiments, we implemented REMIND from scratch, but the original GitHub repository of REMIND helped us significantly (\url{https://github.com/tyler-hayes/REMIND}) \cite{remind}. Furthermore, we relied on Meta Research's FAISS library for PQ (\url{https://github.com/facebookresearch/faiss}) \cite{johnson2019billion}.

\textbf{BIR details:} Brain-inspired replay (BIR) is a generative activation replay method that models the interplay between the Hippocampus and Neocortex during the memory consolidation process of the brain. This approach replays the hidden activations of previously learned data, which are generated by the network's own context-modulated feedback connections.

BIR is designed for class-incremental learning, much like SHARP, and the original paper conducted experiments on datasets similar to ours. As a result, we were able to easily adapt BIR to our experiments, utilizing the official GitHub repository (\url{https://github.com/GMvandeVen/brain-inspired-replay}) \cite{gido2020}.

\textbf{Hyperparameters: } We performed rigorous hyperparameter fine-tuning for all baselines and reported the best performance of each method on the official test sets of all datasets. Our hyperparameter search included any suggested hyperparameters proposed in the original papers.

In terms of optimization, we utilized SGD with momentum for REMIND, ADAM for BIR as recommended in the original implementation, and Adadelta optimizer for SHARP. Although SHARP performs similarly with SGD with momentum if the learning rate is set properly, we observed empirically that Adadelta allowed us to use a single learning rate (1.0) across all experiments. For the sake of simplicity, we retained Adadelta for SHARP. We also tried  Adadelta with a learning rate of 1.0 for REMIND and BIR, but they did not perform better. Please see \textbf{Table}~\ref{tab:bir_sharp_remind_hp} for all hyperparameters.

\begin{table}[ht]
\newcolumntype{C}[1]{>{\centering\arraybackslash}p{#1}}
\centering
\small
\renewcommand{\arraystretch}{1.25}
\caption{Best Hyperparameters for experiment ``Comparison with Activation Replay Methods''. Here we present hyperparameters that are common across all baselines. BS stands for batch size. Training amount is per episode. }
\begin{tabularx}{\textwidth}{Y|C{2.5cm}|Y|Y|Y|Y|}
\hline \hline
 Approach   & Optimizer & Learning rate & BS & Replay BS & Training  \\ 
  \cline{1-6} \multicolumn{6}{c}{Experiment: MNIST $\rightarrow$ EMNIST} \\ \cline{1-6}
SHARP & Adadelta &  1.0 & 512 & 512 & see \textbf{Table}~\ref{tab:sharp_hp_table} \\
REMIND & SGD + Momentum &  0.01  & 512 & 512& 3 epochs \\
BIR & Adam &  0.0001 & 1024 & 1024 & 250 updates \\
  \cline{1-6} \multicolumn{6}{c}{Experiment: CIFAR10 $\rightarrow$ CIFAR100} \\ \cline{1-6}
SHARP & Adadelta &   1.0  & 64 & 64 & see \textbf{Table}~\ref{tab:sharp_hp_table}  \\
REMIND & SGD + Momentum &  0.001  & 64 & 64 & 25 epochs \\
BIR & Adam &  0.0005  & 256 & 256 & 5000 updates  \\
\hline \hline
\end{tabularx}
\label{tab:bir_sharp_remind_hp}
\end{table}

\subsection{Details for Comparison with Raw Replay Methods}
In Section 4.2 of the main paper, we compared SHARP with the following baselines:

\begin{itemize}
    \item \textbf{Experience Replay (ER):} This straightforward replay algorithm enables sequential learning of new classes by augmenting the training batches with samples from all past classes stored in the memory.
    \item \textbf{Dark Experience Replay (DER) :} This method is a simple extension of the ER algorithm \cite{DER}, and it also leverages raw sample replay. The main difference is that, instead of storing hard class labels, it stores the logits of the trained model as targets for replayed samples.
    \item \textbf{Function Distance Regularization (FDR):} This approach also involves replaying raw examples from all previous classes and bears a resemblance to DER. It leverages past examples and network outputs to align current and past outputs, much like DER.
    \item \textbf{iCaRL :} iCaRL is a sophisticated raw replay method that also relies on revisiting all past classes. It is one of the most well-known replay methods and usually perform well across different datasets and architectures. It has three main components: classification by a nearest-mean-of-exemplars rule, prioritized exemplar selection based on herding, and representation learning using knowledge distillation and prototype rehearsal.
    
    \item \textbf{A-GEM :} This method differs from other baseline approaches as it stores raw samples but does not directly use them in training \cite{agem}. Rather, it projects gradients of novel tasks based on the gradients computed for memory samples. Additionally, it requires storing raw examples to compute gradients at specific times during the training process.
\end{itemize}

We relied on \url{https://github.com/aimagelab/mammoth} \cite{DER, boschini2022class} for the implementation of all of these baselines. We tuned all baselines, searching the hyperparameter space for the best performance. Our search included any suggested hyperparameters in \cite{DER}. Also, we tried using both Adadelta and SGD optimizers. Please see \textbf{Table}~\ref{tab:raw_replay_hp} for all hyperaparameters used in experiments.

\begin{table}[ht]
\newcolumntype{C}[1]{>{\centering\arraybackslash}p{#1}}
\centering
\small
\renewcommand{\arraystretch}{1.25}
\caption{Best Hyperparameters for experiment ``Comparison with Raw Replay Methods''. Here we present hyperparameters that are common across all baselines. BS stands for batch size. Training amount is per episode. }
\begin{tabularx}{\textwidth}{Y|C{2.2cm}|C{1.7cm}|Y|Y|C{1.5cm}|C{1.5cm}|}
\hline \hline
 Approach   & Optimizer & Learning rate & BS & Replay BS & Training & Other  \\ 
  \cline{1-7} \multicolumn{7}{c}{Experiment: MNIST} \\ \cline{1-7}
SHARP & Adadelta &  1.0 & 256 & 256 & see \textbf{Table}~\ref{tab:sharp_hp_table} & see \textbf{Table}~\ref{tab:sharp_hp_table} \\
ER & SGD+Momentum &  0.001 & 128 & 128 & 20 & --- \\
DER & SGD+Momentum &  0.001 & 512 & 512 & 20 & $\alpha = 0.2$ \\
FDR & SGD+Momentum&  0.0001 & 256 & 256 & 20 & $\alpha = 0.2$ \\
iCaRL & SGD+Momentum &  0.01 & 128  & 128  & 20 & --- \\
A-GEM & SGD+Momentum &  0.01  & 128 & 128 & 20 & --- \\
 \cline{1-7} \multicolumn{7}{c}{Experiment: FMNIST} \\ \cline{1-7}
 SHARP & Adadelta &  1.0 & 1024 & 1024 & see \textbf{Table}~\ref{tab:sharp_hp_table} & see \textbf{Table}~\ref{tab:sharp_hp_table} \\
ER & SGD+Momentum &  0.0001 & 128 & 128 & 20 & --- \\
DER & SGD+Momentum &  0.001 & 256 & 256 & 20 & $\alpha = 0.2$ \\
FDR & SGD+Momentum &  0.0001 & 128  & 128  & 20 & $\alpha = 0.2$ \\
iCaRL & SGD+Momentum &  0.01 & 128 & 128 & 20 & --- \\
A-GEM & SGD+Momentum &  0.0001  & 256  & 256  & 20 & --- \\
 \cline{1-7} \multicolumn{7}{c}{Experiment: EMNIST} \\ \cline{1-7}
 SHARP & Adadelta &  1.0 & 256 & 256 & see \textbf{Table}~\ref{tab:sharp_hp_table} & see \textbf{Table}~\ref{tab:sharp_hp_table} \\
ER & SGD+Momentum &  0.0001  & 128 & 128 & 20 & --- \\
DER & SGD+Momentum &  0.001 & 512 & 512 & 20 & $\alpha = 0.2$ \\
FDR & SGD+Momentum &  0.001 & 256 & 256 & 20 & $\alpha = 0.2$ \\
iCaRL & SGD+Momentum &  0.01 & 128  & 128  & 20  & --- \\
A-GEM & SGD+Momentum &  0.0001 & 256  & 256  & 20 & --- \\
 \cline{1-7} \multicolumn{7}{c}{Experiment: CIFAR10} \\ \cline{1-7}
 SHARP & Adadelta &  1.0 & 256 & 256 & see \textbf{Table}~\ref{tab:sharp_hp_table} & see \textbf{Table}~\ref{tab:sharp_hp_table} \\
ER & SGD+Momentum &  0.0001 & 256 & 256 & 50 & --- \\
DER & SGD+Momentum &  0.0001 & 256  & 256  & 50 & $\alpha = 0.2$ \\
FDR & SGD+Momentum &  0.0001  & 128 & 128 & 50 & $\alpha = 0.2$ \\
iCaRL & SGD+Momentum &  0.01 & 64 & 64 & 50 & --- \\
A-GEM & SGD+Momentum &  0.001  & 64 & 64 & 50 & --- \\
 \cline{1-7} \multicolumn{7}{c}{Experiment: CIFAR100} \\ \cline{1-7}
  SHARP & Adadelta &  1.0 & 64 & 64 & see \textbf{Table}~\ref{tab:sharp_hp_table} & see \textbf{Table}~\ref{tab:sharp_hp_table} \\
ER & SGD+Momentum &  0.0001  & 256   & 256   & 50 & --- \\
DER & SGD+Momentum &  0.001 & 64 & 64 & 50 & $\alpha = 0.2$ \\
FDR & SGD+Momentum &  0.01 & 64 & 64 & 50 & $\alpha = 0.2$ \\
iCaRL & SGD+Momentum &  0.01  & 64 & 64 & 50 & --- \\
A-GEM & SGD+Momentum &  0.01  & 256 & 256 & 50 & --- \\

\hline \hline
\end{tabularx}
\label{tab:raw_replay_hp}
\end{table}

\subsection{Details for Does SHARP Need Episode Boundaries?}
In Section 4.3 of the main paper, we conducted an evaluation of SHARP in a modified continual learning scenario where episodes are not mutually exclusive in terms of the classes they contain. The experimental details and hyperparameters for these experiments remain the same as in Section 4.2, with one exception: we omitted the reinitialization of rank-0 units. This decision was made because, in this modified scenario, past classes may reappear in the future, and we want to preserve potential forward knowledge transfer by avoiding reinitialization. It is important to note that rank-0 units in the paper are considered idle and do not learn for the task. While this holds true for most rank-0 units, certain units may transition from rank-0 to rank-1 and then be demoted back to rank-0 within a single learning episode. These units will encode some knowledge for the current classes and may contribute to forward knowledge transfer.

\section{Additional Results}

\subsection{Raw Replay Results Across Episodes}
\begin{figure}
  \centering
  \includegraphics[width=\textwidth]{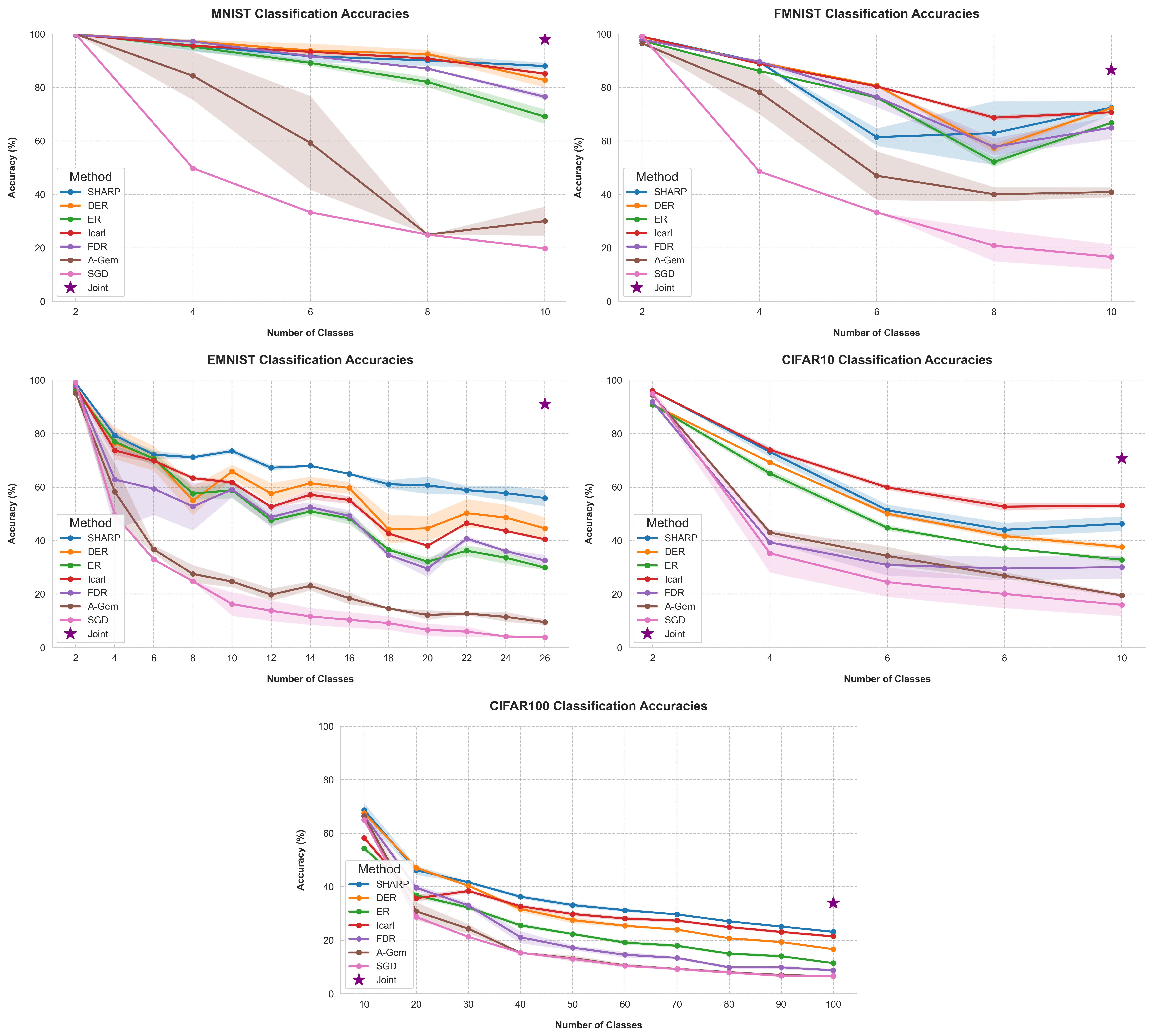}
  \caption{Accuracies on seen classes for MNIST, FMNIST, EMNIST, CIFAR10, and CIFAR100. }
  \label{fig:raw_replay_across_time}
\end{figure}

In Section 4.2 of the main paper, we reported the results in a tabular format, focusing on the final accuracies averaged across all classes. While this metric provides a useful overview of the performance of different methods in continual learning, it does not provide a comprehensive view of the results. To present a more detailed analysis, \textbf{Figure}~\ref{fig:raw_replay_across_time} depicts the accuracies on seen classes as the methods encounter new learning episodes. This visualization offers further insights into the temporal performance of the methods. It is important to note that this is not an independent experiment; rather, we present previously obtained results in a different format to enhance understanding and analysis.

As mentioned in our main paper, SHARP demonstrates the highest average accuracy across all classes, except for the CIFAR10 dataset. However, similar to our findings in Section 4.1 of the main paper, when comparing SHARP to REMIND, SHARP occasionally lags behind other methods in earlier episodes. This can be attributed to the fact that the task becomes progressively more challenging for other baselines as we encounter additional episodes. They are required to uniformly replay all previous classes and distribute limited memory resources across an increasing number of classes. In contrast, SHARP selectively replays past classes and its replay mechanism remains unaffected by the number of episodes seen.

\subsection{Results with Different Memory Budgets}
In the main paper, we presented SHARP's performance based on a specific fixed memory budget, since we are constrained by page limits. However, for the sake of completeness, we now provide additional results in \textbf{Table}~\ref{tab:classification_accuracies_half} and \textbf{Table}~\ref{tab:classification_accuracies_double}, which correspond to scenarios where the memory budgets are halved and doubled, respectively. Please note that we have chosen to present results only for SHARP, iCaRL, and DER in this analysis. This decision is based on the considerable accuracy gap observed between SHARP and the other methods. Given this significant gap, we do not anticipate that modifying the memory budget will substantially narrow or close the performance difference.

As anticipated, when the memory budget is halved, the performance of all methods tends to decrease. However, in general, SHARP demonstrates more resilience in this low-memory regime. For instance, on the MNIST dataset, SHARP's performance only decreased by 2.7\%, whereas the next best-performing method, iCaRL, suffered a drop of 11.7\%. Similarly, on FMNIST, SHARP's performance dropped by 3.5\%, while iCaRL experienced a drop of 7\%. This overall trend can be attributed to the fact that other baselines now have very limited capacity to store samples per class in order to uniformly replay all previously encountered classes.

It is important to note that there are exceptions to this general trend. For instance, iCaRL demonstrates better preservation of performance on the CIFAR10 dataset. We believe that iCaRL's robust performance on datasets where samples within the same class exhibit significant variations can be attributed to its non-random sample selection strategy. This observation suggests that SHARP's random selection strategy could potentially be replaced with a more sophisticated selection strategy to further enhance its performance.

Conversely, when the memory budget is doubled, all methods exhibit an improvement in accuracy. However, in this scenario, the raw replay methods, DER and iCaRL, benefit more prominently from the increased memory budget. Additionally, SHARP no longer maintains its lead on the FMNIST dataset. This outcome is expected because in a high-memory regime, replaying raw examples of all previously encountered classes approximates the use of i.i.d. data, and the raw replay methods become closer to joint training, which represents the upper bound performance for continual learning.

\begin{table}[ht]
\newcolumntype{C}[1]{>{\centering\arraybackslash}p{#1}}
\centering
\renewcommand{\arraystretch}{1.25}
\caption{Half Memory Budget. Average accuracy across classes after all episodes. MNIST/FMNIST/EMNIST memory: 0.0203 MB (26 images), CIFAR10 memory: 0.8192 MB ($\approx267$ images), CIFAR100 memory: 2.048 MB ($\approx667$ images).}
\begin{tabularx}{0.9\textwidth}{Y|Y|Y|Y|Y|Y}
\hline \hline
\textbf{Method}    & \textbf{MNIST} & \textbf{FMNIST} & \textbf{EMNIST} & \textbf{CIFAR10} & \textbf{CIFAR100} \\ \hline \hline
Joint & 98.0 ($\pm 0.0$) & 86.6 ($\pm 1.0$) & 91.0 ($\pm 0.3$) & 70.8 ($\pm 0.4$) & 33.9 ($\pm 0.8$) \\ \hline
\textbf{SHARP}  & \textbf{85.3} ($\pm 0.8$) & \textbf{69.0} ($\pm 2.5$) & \textbf{55.8} ($\pm 1.3$) &  41.3 ($\pm 1.0$) & \textbf{21.5} ($\pm 1.5$) \\
iCaRL  & 73.4 ($\pm 1.3$) & 63.7 ($\pm 1.0$) & 55.1 ($\pm 0.8$) &  \textbf{52.6} ($\pm 0.9$) & 19.6 ($\pm 0.4$) \\
DER  & 73.2 ($\pm 1.9$) & 64.5 ($\pm 3.1$) & 34.5 ($\pm 2.0$) &  32.6 ($\pm 1.2$) & 11.4 ($\pm 0.4$) \\
\hline
SGD               & 19.8 ($\pm 0.1$)            & 16.6 ($\pm 4.7$)            & 3.9 ($\pm 0.0$)             & 16.0 ($\pm 4.2$)              & 6.6 ($\pm 0.1$)               \\

\hline \hline
\end{tabularx}
\label{tab:classification_accuracies_half}
\end{table}

\begin{table}[ht]
\newcolumntype{C}[1]{>{\centering\arraybackslash}p{#1}}
\centering
\renewcommand{\arraystretch}{1.25}
\caption{Double Memory Budget. Average accuracy across classes after all episodes. MNIST/FMNIST/EMNIST memory: 0.0784 MB (100 images), CIFAR10 memory: 3.2768 MB ($\approx1068 $ images), CIFAR100 memory: 8.192 MB ($\approx 2668$ images).}
\begin{tabularx}{0.9\textwidth}{Y|Y|Y|Y|Y|Y}
\hline \hline
\textbf{Method}    & \textbf{MNIST} & \textbf{FMNIST} & \textbf{EMNIST} & \textbf{CIFAR10} & \textbf{CIFAR100} \\ \hline \hline
Joint & 98.0 ($\pm 0.0$) & 86.6 ($\pm 1.0$) & 91.0 ($\pm 0.3$) & 70.8 ($\pm 0.4$) & 33.9 ($\pm 0.8$) \\ \hline
\textbf{SHARP}  & \textbf{91.7} ($\pm 0.9$) & 72.8 ($\pm 2.0$) & \textbf{61.1} ($\pm 1.6$) &  49.3 ($\pm 0.1$) & \textbf{23.0} ($\pm 0.7$) \\
iCaRL  & 86.0 ($\pm 0.5$) & 71.0 ($\pm 0.7$) & 59.8 ($\pm 1.2$) &  \textbf{53.6} ($\pm 1.0$) & 22.9 ($\pm 0.6$) \\
DER  & 87.3 ($\pm 1.4$) & \textbf{75.6} ($\pm 0.5$) & 51.3 ($\pm 2.7$) &  41.1 ($\pm 0.9$) & 20.7 ($\pm 0.6$) \\
\hline
SGD               & 19.8 ($\pm 0.1$)            & 16.6 ($\pm 4.7$)            & 3.9 ($\pm 0.0$)             & 16.0 ($\pm 4.2$)              & 6.6 ($\pm 0.1$)               \\

\hline \hline
\end{tabularx}
\label{tab:classification_accuracies_double}
\end{table}

\subsection{Experiments with Various STM Temporal Window Sizes (i.e., $m$)}
Throughout all the conducted experiments, SHARP employed the replay of activations from the previous episode. This was achieved by setting the short-term memory parameter, denoted as $m$, to 1. However, it is important to note that our algorithm is flexible and can function with any value of $m$, even allowing for scenarios where $m$ is set to 0  (no replay). Note that in the case of $m=0$, we temporarily store activations from the current episode to pass them to long-term memory, but we do not perform any replay.

In this section, we conduct additional experiments with different values of $m$ to explore its impact on SHARP's performance. It is important to note that the memory budget is fixed to 50 activations, which is consistent with our main experiments, regardless of the value of $m$. Consequently, as $m$ increases, SHARP is required to allocate the same fixed budget across an increasing number of classes.

\begin{figure}
  \centering
  \includegraphics[width=0.7\textwidth]{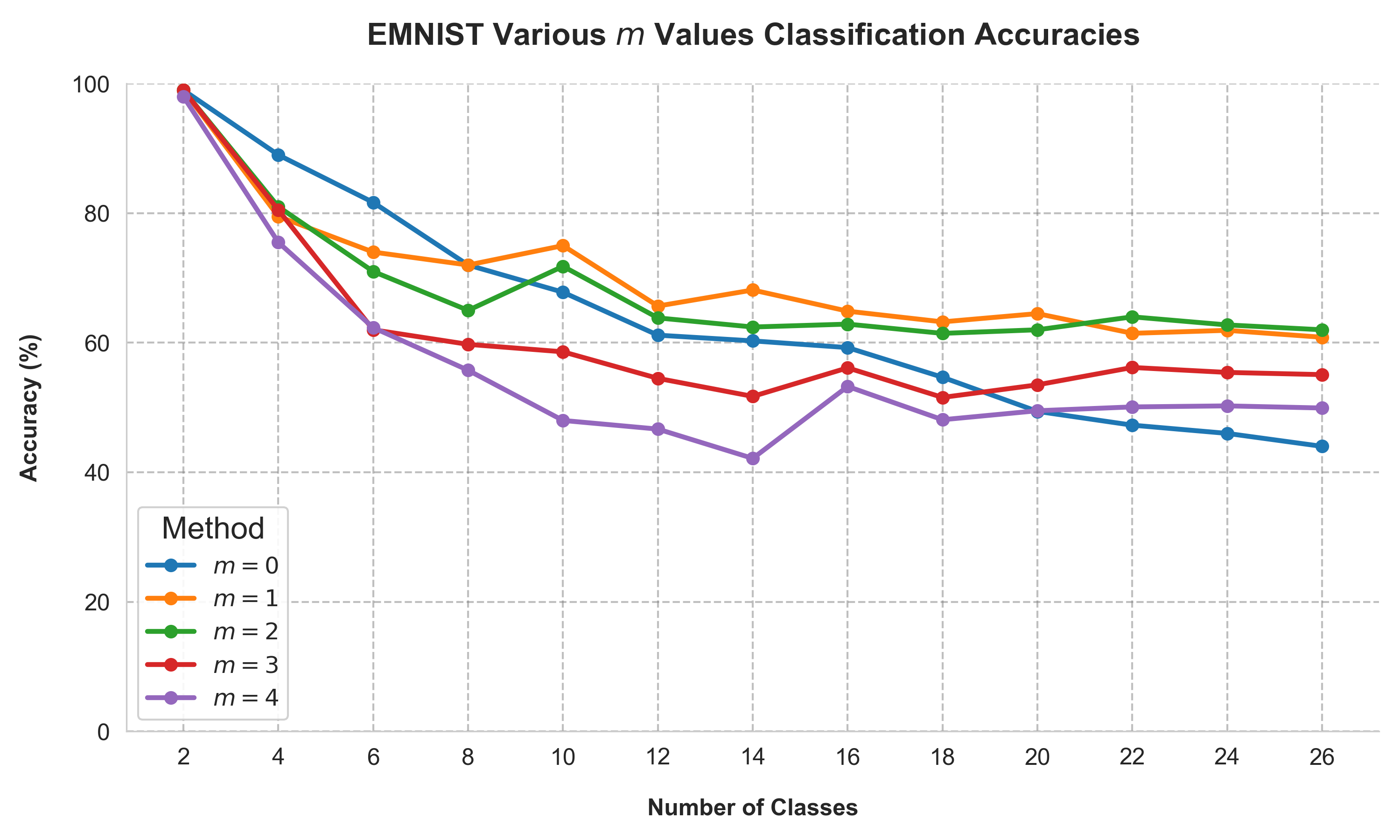}
  \caption{Accuracies on seen classes on EMNIST for various $m$ values.}
  \label{fig:EMNIST_values_m_classification_accuracies}
\end{figure}

\textbf{Figure}~\ref{fig:EMNIST_values_m_classification_accuracies} showcases the results obtained for various values of $m$ (0, 1, 2, 3, 4) on the EMNIST dataset. Remarkably, even without any replay (i.e., $m=0$), SHARP achieves non-trivial performance. Notably, it even outperforms the others in the second and third episodes. However, as subsequent episodes unfold, its performance experiences a rapid decline, causing SHARP without replay to lag behind the replay versions. This observation emphasizes that the strengths of SHARP are not solely attributed to replay. Instead, architectural components like the ranking system and connection rewiring play significant roles in retaining knowledge over time.

Conversely, when $m = 2$, we observe a slightly improved performance towards the end of the learning trajectory compared to when $m = 1$. However, as we increase $m$ to values higher than 2, there is a noticeable decrease in the overall performance. This decline can be attributed to the fact that SHARP is now required to rely on a very limited number of samples per class during replay, which can potentially lead to overfitting. Therefore, it is preferable to prioritize the replay of recently seen classes when storing a large number of samples per class is not feasible. This observation aligns with insights from neuroscience as well. The brain also faces a similar trade-off, as computational efficiency is crucial for survival. Hence, the brain prioritizes the replay of recently experienced events rather than revisiting all past experiences.

\section{SHARP's Implementation and Compute Resources}
For all experiments, we conducted them on Ubuntu 20.04 using a NVIDIA GeForce RTX 3090 GPU with CUDA 12.0. We utilized PyTorch 1.13.1 and Python 3.10.9 for our implementation. Finally, we relied on ``Avalanche: an End-to-End Library for Continual Learning'' for creating continual learning scenarios \cite{lomonaco2021avalanche}.


\end{document}